\documentclass[lettersize,journal]{IEEEtran}
\usepackage{amsmath,amsfonts}
\usepackage{array}
\usepackage[caption=false,font=normalsize,labelfont=sf,textfont=sf]{subfig}
\usepackage{textcomp}
\usepackage{stfloats}
\usepackage{url}
\usepackage{verbatim}
\usepackage{graphicx}
\usepackage{amssymb}
\usepackage{color} 
\usepackage{threeparttable}
\usepackage{amsthm,amsmath,amssymb}
\usepackage{mathrsfs}
\usepackage{indentfirst}
\usepackage{algorithm} 
\usepackage{makecell}
\usepackage{algpseudocode}
\usepackage{multirow}
\usepackage{url}
\usepackage{footnote}
\usepackage{cite}
\usepackage{array}
\usepackage{verbatim}
\usepackage{colortbl}
\usepackage{booktabs}

\newcommand{\tabincell}[2]{\begin{tabular}{@{}#1@{}}#2\end{tabular}}
\usepackage{balance}
\begin{document}
\title{Joint Attention-Guided Feature Fusion Network for Saliency Detection of Surface Defects}
\author{Xiaoheng Jiang, Feng Yan, Yang Lu, Ke Wang, Shuai Guo\\
Tianzhu Zhang, Yanwei Pang, \emph{Senior Member, IEEE}, Jianwei Niu, \emph{Senior Member, IEEE}, and Mingliang Xu
\thanks{This work was supported in part by National Key R\&D Program of China under Grant 2021YFB3301504, in part by the National Natural Science Foundation of China under Grant 62172371, U21B2037, 62036010, 62102370, 61903341, 62106232, in part by China Postdoctoral Science Foundation under Grant 2021TQ0301, and in part by Foundation for University Key Research of Henan Province (21A520040, 21A520002), Hangzhou Innovation Institute, Beihang University (NO. 2020-Y4-A-020), and CAAI-Huawei MindSpore OpenFund. (\emph{Corresponding author: Jianwei Niu, Mingliang Xu}.)

Xiaoheng Jiang, Yang Lu, Ke Wang, Shuai Guo, and Mingliang Xu are with the School of Computer and Artificial Intelligence, Zhengzhou University, Engineering Research Center of Intelligent Swarm Systems, Ministry of Education, National Supercomputing Center in Zhengzhou, Zhengzhou 450001, China (e-mail: jiangxiaoheng@zzu.edu.cn;  ieylu@zzu.edu.cn; iekwang@zzu.edu.cn; iesguo@zzu.edu.cn; iexumingliang@zzu.edu.cn).

Feng Yan is with the School of Computer and Artificial Intelligence, Zhengzhou University, Zhengzhou 450001, China (ieyanfeng@163.com).

Tianzhu Zhang is with the School of Information Science and Technology, University of Science and Technology of China, Hefei 230026, China (e-mail: tzzhang@ustc.edu.cn).

Yanwei Pang is with the School of Electrical Automation and Information Engineering, Tianjin University, Tianjin 300072, China (e-mail: pyw@tju.edu.cn).

Jianwei Niu is with the State Key Laboratory of Virtual Reality Technology and Systems, School of Computer Science and Engineering, Beihang University, Beijing 100191, China, and Hangzhou Innovation Institute of Beihang University, Hangzhou 310051, China (e-mail: niujianwei@buaa.edu.cn).
}}
\markboth{IEEE TRANSACTIONS ON INSTRUMENTATION AND MEASUREMENT,~Vol.~xx, No.~xx, xx~2022}%
{How to Use the IEEEtran \LaTeX \ Templates}

\maketitle
\begin{abstract}
Surface defect inspection plays an important role in the process of industrial manufacture and production. Though Convolutional Neural Network (CNN) based defect inspection methods have made huge leaps, they still confront a lot of challenges such as defect scale variation, complex background, low contrast, and so on.
To address these issues, we propose a joint attention-guided feature fusion network (JAFFNet) for saliency detection of surface defects based on the encoder-decoder network.
JAFFNet mainly incorporates a joint attention-guided feature fusion (JAFF) module into decoding stages to adaptively fuse low-level and high-level features.
The JAFF module learns to emphasize defect features and suppress background noise during feature fusion, which is beneficial for detecting low-contrast defects.
In addition, JAFFNet introduces a dense receptive field (DRF) module following the encoder to capture features with rich context information, which helps detect defects of different scales.
The JAFF module mainly utilizes a learned joint channel-spatial attention map provided by high-level semantic features to guide feature fusion. 
The attention map makes the model pay more attention to defect features.
The DRF module utilizes a sequence of multi-receptive-field (MRF) units with each taking as inputs all the preceding MRF feature maps and the original input. The obtained DRF features capture rich context information with a large range of receptive fields.
Extensive experiments conducted on SD-saliency-900, Magnetic tile, and DAGM 2007 indicate that our method achieves promising performance in comparison with other state-of-the-art methods. Meanwhile, our method reaches a real-time defect detection speed of 66 FPS.
\end{abstract}


\section{Introduction}
\IEEEPARstart{S}{urface} defect inspection is a key task in the process of industrial production and is essential for product quality control. Compared with the manual defect inspection methods, computer vision based automatic defect inspection technologies have become more popular in industrial production due to their superior defect inspection performance with faster speed and higher accuracy.

\begin{figure}[t]
	\centering
		\includegraphics[scale=0.6]{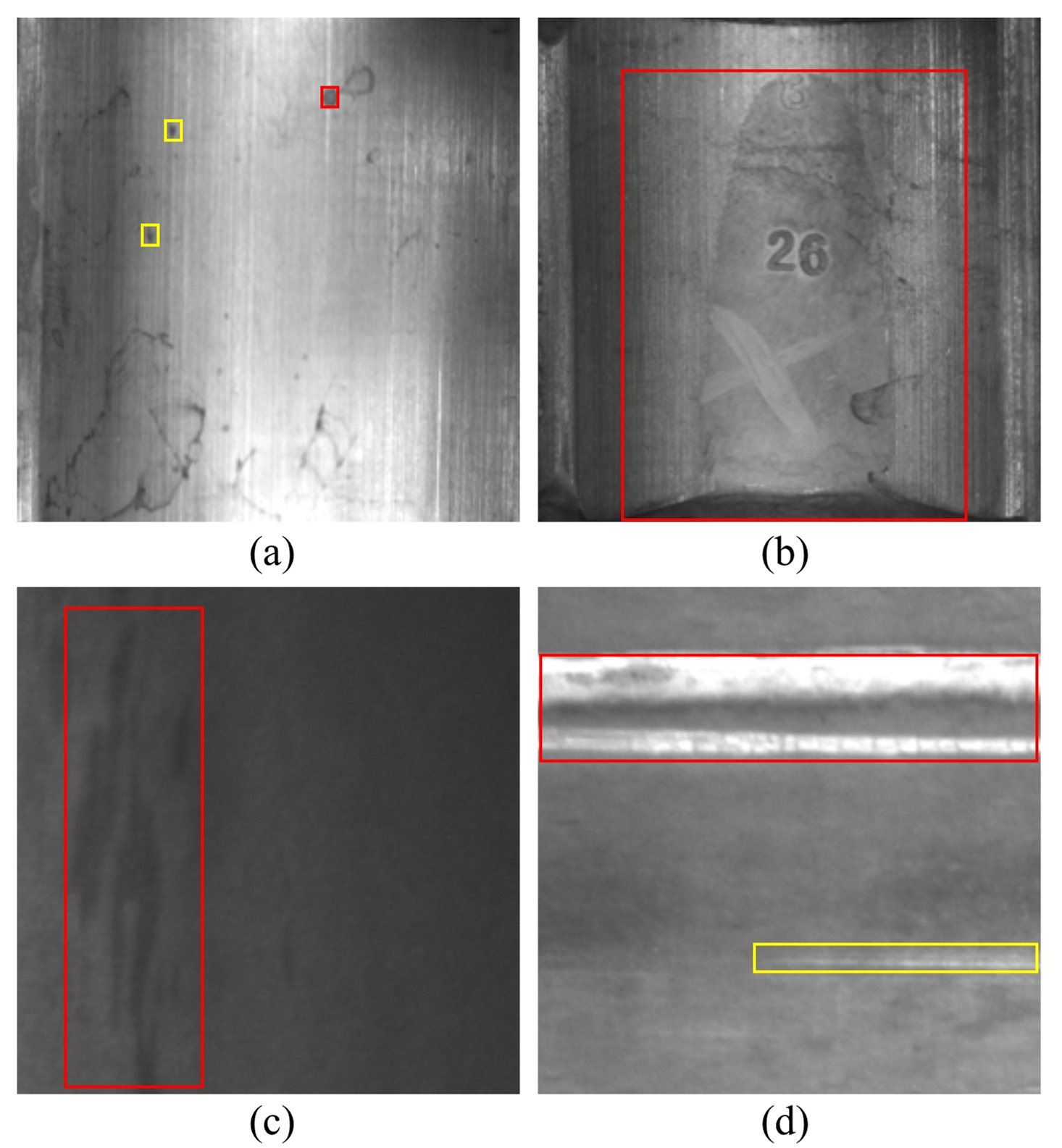}
	\caption{Challenges of surface defect inspection. (a) and (b) defects with different scales. (c) defects with low contrast. (a) and (d) interference factors in the background. The defects and interference factors are represented by red and yellow rectangles, respectively.}
	\label{defects}
\end{figure}

Traditional defect inspection approaches generally consider the surface defect inspection as a texture analysis issue and exploit several classic strategies such as texture filters \cite{Jiang2020Gabor}, texture statistics \cite{Pastor2019GLCM,Luo2019LBP}, texture modeling \cite{hanmandlu2015detection}, and texture structure \cite{Berwo2021Canny}.
These methods rely heavily on specific texture information and work well when the defects are simple. However, the surface defects in real industrial scenes usually exhibit complexity and diversity in appearance and scale, which brings a huge challenge for accurate defect inspection. 
Fig.~\ref{defects} demonstrates several typical issues about surface defect inspection, including scale variation, low contrast, and background interference. The red and yellow rectangles in Fig. \ref{defects} represent the defects and background interference, respectively. 
Fig.~\ref{defects} (a) and (b) show that the defects vary largely in scale. Some surface defects are very small and have less than 80 pixels in a 256 $\times$ 256 image, as shown in Fig.~\ref{defects} (a).
Fig.~\ref{defects} (c) shows the low contrast between the defects and the background caused by inappropriate lighting conditions.
Fig.~\ref{defects} (a) and (d) show that there exist some interference factors in the background which are very similar to the defects and are hard to discriminate. 

Recently, deep learning methods based on convolutional neural network (CNN) have made great progress in many computer vision tasks such as image classification \cite{He2016Deep, Huang2017Densely, gao2019res2net}, object detection \cite{Sun2021Sparse, Chen2021YOLOF, KimL2020PAA}, image segmentation\cite{Sun2021C2FNet, 21Sal100K, Xie2022PGNet}, and so on. These methods are designed for general objects and can not directly generalize to surface defect inspection due to the above mentioned challenges.   
To handle these challenges, researchers have designed CNN-based models that target surface defect inspection, such as the region-level defect inspection methods \cite{He2020An, Wei2020Detecting, Cui2021SDD, Su2021RCAG-Net, Tu2021An} and the pixel-level ones \cite{Huang2020Compact, Zhang2021MCnet, Li2021An, Song2020EDRNet, Zhou2022DACNet}. 
Among these works, the pixel-level methods can provide more detailed information about defects, such as boundary, shape, and size. 
Most of these methods adopt the encoder-decoder structure as the basic backbone, in which the decoder can be regarded as the fusion process of high-level features from the top layers and low-level features from the corresponding bottom layers. The high-level features contain more abstract semantic information, while the low-level features contain more fine details. The combination of the two-level features is beneficial to defect inspection. 
However, these methods still suffer from a certain amount of inspection errors when the defects show weak appearances, which are usually characterized by low contrast, small area, or subtle scratch. 
That is mainly because these methods like \cite{Zhou2022DACNet, Zhang2021MCnet, Li2021An} simply adopt direct addition or concatenation operations to combine low-level and high-level features, in which the features related to defects are prone to be drowned by the background during feature fusion.

To solve this problem, we present a joint attention-guided feature fusion (JAFF) module, which can adaptively reserve features of defects during feature fusion.
JAFF first computes a channel-spatial attention map using the high-level features and then uses it to refine the corresponding low-level features. Finally, JAFF concatenates the refined low-level features and high-level features in channel dimension. 
The high-level features are used to generate the attention map based on the fact that they contain rich semantic information about the defects. 
As a result, the obtained attention map can emphasize the most meaningful low-level defect features and suppress background noise during feature fusion, resulting in robust defect-focused features. 
In addition, context information is also crucial in defect detection, especially for those defects of various scales. As the scale of defects changes significantly, the size of the receptive field should change accordingly. 
To handle this problem, we present a dense receptive field (DRF) module to capture rich local context information with dense receptive fields, as well as global context information. DRF utilizes the multi-receptive-field (MRF) units connected densely to promote the multi-scale representation ability of features. 

Based on the proposed JAFF module and DRF module, we develop the joint attention-guided feature fusion network (JAFFNet) for saliency detection of surface defects. 
In summary, the main contributions are as follows:
\begin{enumerate}
\itemsep=0pt
\item We develop a joint attention-guided feature fusion network (JAFFNet) for saliency detection of surface defects by introducing two plug-and-play modules, which can achieve end-to-end defect detection. 
\item We present a joint attention-guided feature fusion (JAFF) module to effectively fuse high-level and low-level features. It is able to select valuable low-level defect features and suppress background interference during feature fusion via the learned joint channel-spatial attention map.
\item We present a dense receptive field (DRF) module to capture context information with larger and denser scale ranges. It exploits rich context information by densely connecting a series of multi-receptive-field units and can handle defects with various scales. 
\item The experiments on three publicly available surface defect datasets, including SD-saliency-900, DAGM 2007, and Magnetic Tile, demonstrate that the proposed method not only achieves promising defect detection performance but also reaches a real-time detection speed of 66 FPS.
\end{enumerate}
\begin{figure*}[tbh]
	\centering
		\includegraphics[scale=0.56]{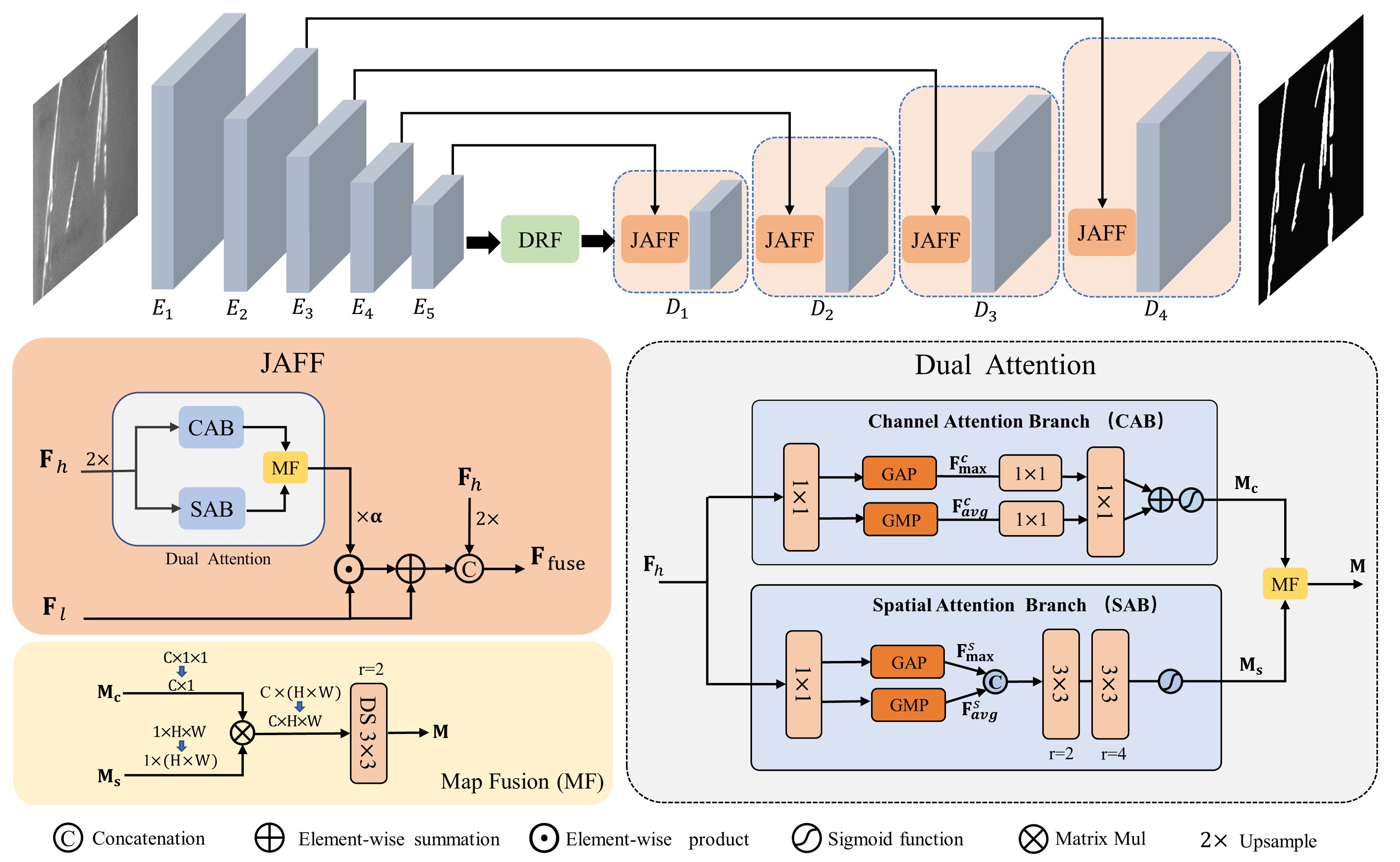}
	\caption{Architecture of the proposed network. Our model consists of an encoder and a decoder, where we obtain multi-level features with channels 64, 128, 256, 512, and 512 from five encoding stages $ E_1 \sim E_5 $, respectively. And $ D_1 \sim D_4 $ represent four decoding stages with each including a joint attention-guided feature fusion (JAFF) module and a convolution block. And the JAFF focuses on the fusion of high-level and low-level features. It incorporates a dual attention module consisting of a channel attention branch (CAB) and a spatial attention branch (SAB) to generate the learned channel-spatial attention map that provides guidance for feature fusion. The dense receptive field (DRF) module after the encoder is used to capture dense context information. And the ``DS'' and ``r'' denote depthwise separable convolution and rate of dilated convolution, respectively.}
	\label{network}
\end{figure*}

\section{Related Works}
\subsection{Traditional defect inspection methods}
Most traditional surface defect inspection methods are based on texture analysis, which can be broadly classified into four categories: filter-based, statistic-based, model-based, and structure-based approaches.
Specifically, the filter-based methods analyze texture features through filters, such as Fourier transform, Gabor transform \cite{Jiang2020Gabor}, and Wavelet transform.
The statistic-based methods analyze texture features through statistical distribution characteristics of the image, such as gray-level co-occurrence matrix (GLCM) \cite{Pastor2019GLCM}, local binary pattern (LBP) \cite{Luo2019LBP}.
The model-based methods describe texture features through statistics of model parameters, such as the random field model, and fractal model \cite{hanmandlu2015detection}.
The structure-based approaches analyze texture features through texture primitives and spatial placement rules, such as \cite{Berwo2021Canny}.
And these methods are most customized for specific types of defects, with poor reusability and generalization ability. In addition, these methods cannot effectively deal with complicated defects.

\subsection{Deep-learning-based defect inspection methods}
Compared with traditional surface defect inspection methods, deep-learning-based defect inspection methods are able to handle defects with weak characteristics and complex background, and show superiority in complex scenes. And we broadly divide these methods into two categories: region-level and pixel-level defect inspection methods.

\subsubsection{Region-level inspection methods} These methods locate defects by bounding boxes. To improve the defect detection ability of the model,
He et al. \cite{He2020An} first integrate multi-level features into one feature and then feed it into the region proposal network to generate high-quality defect proposals.
Wei et al. \cite{Wei2020Detecting} incorporate the attention-related visual gain mechanism into the Faster RCNN model to improve the discrimination ability of small defects.
However, these detectors obtain bounding boxes based on region proposals, with high accuracy but slow speed.
Therefore, Cui et al. \cite{Cui2021SDD} design a fast and accurate detector called SDDNet, which detects small defects by passing fine-grained details of low-level features to all deep features.  
Su et al. \cite{Su2021RCAG-Net} adopt a novel attention module (RCAG) to fuse multi-scale features, with the aim of emphasizing defect features and suppressing background noise.
Different from  \cite{Cui2021SDD} and \cite{Su2021RCAG-Net}, Tu et al. \cite{Tu2021An} achieve accurate defect detection by adopting CIoU loss and introducing the Gaussian function to estimate the coordinates of the prediction boxes.

\subsubsection{Pixel-level inspection methods} These methods can provide more structural details of defects than region-level methods, such as boundary, shape, and size. It is essential for accurate defect detection to capture and integrate multiple context information effectively. To this end,
Huang et al. \cite{Huang2020Compact} apply an atrous spatial pyramid pooling (ASPP) in the proposed lightweight defect segmentation network to capture multiple context information.
Zhang et al. \cite{Zhang2021MCnet} integrate multiple context information through the pyramid pooling module (PPM) and attention module, with the aim of enhancing defect features and filtering out noise.
Li et al. \cite{Li2021An} integrate multi-scale features from encoder blocks step-by-step, which sequentially fuses two adjacent scale features and three adjacent scale features.
In addition, the attention mechanism is also often used in defect segmentation to address those defects with complex background. For example,
Song et al. \cite{Song2020EDRNet} incorporate the attention mechanism into the model to steer it to focus more on defect features.
Zhou et al. \cite{Zhou2022DACNet} introduce dense attention cues into the decoder to make it more defect-focused.

Different from these existing methods, we propose a joint attention-guided feature fusion network for saliency detection of surface defects. Specifically, we design two modules to improve the defect detection performance of encoder-decoder architecture. One is called JAFF module, which is used at each decoding stage to retain more defect features during the fusion of low-level and high-level features. The other is called DRF module, which is embedded after the fifth encoding stage to capture dense context information and strengthen the representation ability of deep features. And the proposed two modules greatly improve defect detection performance of the network in complex scenes.

\subsection{Attention Mechanism in CNNs}
The attention mechanism can selectively focus on important information while ignoring less useful information, which is important for understanding complex scenes.
Hu et al. \cite{hu2018squeeze} first propose channel attention and perform adaptive feature recalibration by explicitly modeling global information. 
Due to the limitation of channel attention, Woo et al. \cite{woo2018cbam} propose the convolutional block attention module (CBAM) which applies both channel-wise and spatial attention in sequential.
CBAM not only introduces spatial attention but also introduces both max-pooled and average-pooled features in the spatial axis into channel attention.
Park et al. \cite{park2018bam} also design the bottleneck attention module (BAM) that combines channel-wise \cite{hu2018squeeze} and spatial attention branches through element-wise summation.
BAM computes spatial attention by learnable weighted channel pooling.
Different from these methods, the introduced dual attention module calculates the channel-spatial attention map by performing an element-wise product between each channel attention weight and all spatial attention weights. 
The joint weights generated by the element-wise product are used to guide feature fusion, which can significantly magnify the difference between defect features and background disturbances during feature fusion.
Furthermore, we apply dilated convolution in the spatial dimension to enlarge receptive fields, which can determine where to focus from a broad view. 

\section{Method}
\subsection{Network Architecture Overview}
As illustrated in Fig.~\ref{network}, our network is composed of an encoder and a decoder. The encoder contains five encoding stages $ E_1\sim E_5 $. Among them, $ E_1\sim E_4 $ are adopted from ResNet18 \cite{He2016Deep}, where the input layer is replaced by a 3 $\times$ 3 convolution with stride of 1 and its subsequent pooling operation is also removed. 
This enables the encoder to obtain feature maps with higher resolution at early stages but decreases its overall receptive field.
Thus, we add an extra stage $E_5$ to enlarge the receptive field after stage $E_4$, which is composed of a 2 $\times$ 2 max-pooling operation and two basic res-blocks. After the encoder, we employ the proposed dense receptive field (DRF) module to capture rich context information. The decoder contains four decoding stages $ D_1 \sim D_4 $. Each decoding stage applies the proposed joint attention-guided feature fusion (JAFF) module for feature fusion and a 3 $\times$ 3 convolution block for channel reduction.

\subsection{Joint Attention-guided Feature Fusion Module}
As is known to all, low-level features at the bottom layer contain more details, such as edge, shape, and contour, which are as important as high-level abstract semantics at the top layers. 
The combination of two-level complementary information is beneficial to generate more accurate and complete saliency maps. But because of weak defects and background noise, direct concatenation of two-level features may make the defect features submerged in the background. 
Thus, we propose a joint attention-guided feature fusion (JAFF) module to guide the fusion of low-level and high-level features. 
As illustrated in Fig.~\ref{network}, the JAFF module takes as the input the high-level features from the previous decoding stage and the corresponding low-level features from the encoding stage. The JAFF module adaptively weights low-level features through the learned channel-spatial attention map during feature fusion when dealing with weak defects. The attention map can selectively emphasize important defect features and suppress background noise in both channel and spatial dimensions, making the model pay more attention to defect regions.
\subsubsection{Dual Attention Module}

Given high-level features $\mathbf{F}_h$, the dual attention module described in Fig.~\ref{network} generates the joint channel-spatial attention map through the following steps. First, considering the feature misalignment between low-level features $\mathbf{F}_l$ and $\mathbf{F}_h$, the $\mathbf{F}_h$ is upsampled to the same resolution as $\mathbf{F}_l$ and passed through the channel attention branch and spatial attention branch respectively, obtaining channel attention map $\mathbf{M_c}$ and spatial attention map $\mathbf{M_s}$. Then two maps are fused into one map in a simple way. The details of two attention branches and the fusion details of two attention maps are described as follows.

\textbf{Channel Attention Branch.} Given the upsampled high-level features $\mathbf{F}_h$, it aims to produce a 1D channel attention map that dynamically selects important features in the channel dimension.
Specifically, it first performs a 1 $\times$ 1 convolution on features $\mathbf{F}_h$ and then employs global average pooling (GAP) and global max-pooling (GMP) operations in spatial dimension to compress spatial information, obtaining average-pooled features and max-pooled features, denoted as $\mathbf{F}_{avg}^c$ and $\mathbf{F}_{max}^c$, respectively. 
The pooled features are then fed into a $1\times1$ convolution respectively, followed by a shared $1\times1$ convolution, to capture global dependencies and generate channel weights.
Finally, the resulting channel weights are added together, obtaining channel attention map $\mathbf{M_c}\in\mathbb{R}^{C\times 1\times 1}$ through a sigmoid function. Mathematically, we have
\begin{equation}
\label{deqn_ex1a}
\begin{split}
\mathbf{M_c}=\sigma(\mathcal{F}_3^{1\times1}(\mathcal{F}_1^{1\times1}(\mathbf{F}_{avg}^c))+
\mathcal{F}_3^{1\times1}({\mathcal{F}_2^{1\times1}(\mathbf{F}}_{max}^c)))
\end{split}
\end{equation}
where $\sigma $ denotes the sigmoid function, $\mathcal{F}_1^{1\times1}$, $\mathcal{F}_2^{1\times1}$ and $\mathcal{F}_3^{1\times1}$ denote three $1\times1$ convolutions, respectively.

\textbf{Spatial Attention Branch.} With the upsampled high-level features $\mathbf{F}_h$, it generates a 2D spatial attention map that adaptively emphasizes important features in the spatial dimension.
Likewise, it first performs a 1 $\times$ 1 convolution on features $\mathbf{F}_h$ and then applies GAP and GMP operations in the channel dimension to compress channel information, generating average-pooled features and max-pooled features, denoted as $\mathbf{F}_{avg}^s$ and $\mathbf{F}_{max}^s$, respectively. 
Both pooled features are concatenated together, followed by a sub-network to generate spatial weights with a larger receptive field. And the sub-network is composed of two $3\times3$ dilated convolutions with rate = 2 and rate = 4 respectively. Finally, the resulting weights are through a sigmoid function, generating spatial attention map $\mathbf{M_s}\in\mathbb{R}^{1 \times H \times W}$. Mathematically, we have
\begin{equation}
\label{deqn_ex1a}
\mathbf{M_s}=\sigma(\mathcal{F}^{3\times3, 4}(\mathcal{F}^{3\times3, 2}(CAT(\mathbf{F}_{\max}^s, \mathbf{F}_{avg}^s))))
\end{equation}
where ${\mathcal{F}}^{3\times3, 2}$ and ${\mathcal{F}}^{3\times3, 4}$ represent two $3\times3$ dilated convolutions with rate = 2 and rate = 4 respectively, and $CAT$ denotes concatenation operation.

\textbf{Map Fusion.} With the attention maps $\mathbf{M_c}$ and $\mathbf{M_s}$ obtained from two branches, we fuse them through the following steps. 
The $\mathbf{M_c}$ and $\mathbf{M_s}$ are reshaped into $\mathbb{R}^{C \times 1}$ and $\mathbb{R}^{1 \times N}$ respectively, where $N = H \times W$, followed by a matrix multiplication operation. The result is reshaped into $\mathbb{R}^{C \times H \times W }$ and fed into a depthwise separable 3 $\times$ 3 dilated convolution with rate = 2 to model spatial dependencies, generating the final channel-spatial map $\mathbf{M}$. Formally, we have
\begin{equation}
\label{deqn_ex1a}
\mathbf{M}= {\mathcal{D}}^{3\times3, 2}({\mathbf{M_c}} \otimes {\mathbf{M_s}})
\end{equation}
where $\mathcal{D}^{3\times3, 2}$ represents a depthwise separable $3\times3$ dilated convolution with rate = 2, and $\otimes$ denotes matrix multiplication.  
\subsubsection {Joint attention-guided feature fusion} 
With the joint channel-spatial attention map $\mathbf{M}$ obtained, it is performed an element-wise product with the corresponding  $\mathbf{F}_l$. The result is then multiplied by a scale parameter $\mathbf{\alpha}$, followed by an element-wise sum operation with the $\mathbf{F}_l$, obtaining the refined low-level features $\mathbf{F}_{l}^{\prime}$. Finally, $\mathbf{F}_{l}^{\prime}$ is concatenated with the upsampled features $\mathbf{F}_h$ to produce the fused features $\mathbf{F}_{\rm fuse}$ in the channel dimension. In short, we have
\begin{equation}
\label{deqn_ex1a}
\mathbf{F}_{l}^{\prime}=\mathbf{\alpha} \cdot \mathbf{F}_{l} \odot \mathbf{M} + \mathbf{F}_{l}
\end{equation}
\begin{equation}
\label{deqn_ex1a}
\mathbf{F}_{\rm fuse} =CAT( \mathbf{F}_{l}^{\prime}, \mathbf{F}_{h} )
\end{equation}
where $\mathbf{\alpha}$ is a learnable parameter, which can adaptively model relationship between the $\mathbf{F}_l$ and $\mathbf{F}_{l}^{\prime}$. 
\subsection{Dense Receptive Field Module}

Different from general objects, defects vary widely in scale. To handle this case, the extracted features require covering a diverse range of receptive fields. For this goal, we present a dense receptive field (DRF) module to capture context information with dense and large scales.

\begin{figure}[t]
	\centering	 
        \includegraphics[width=3.3in,height=2.1in]{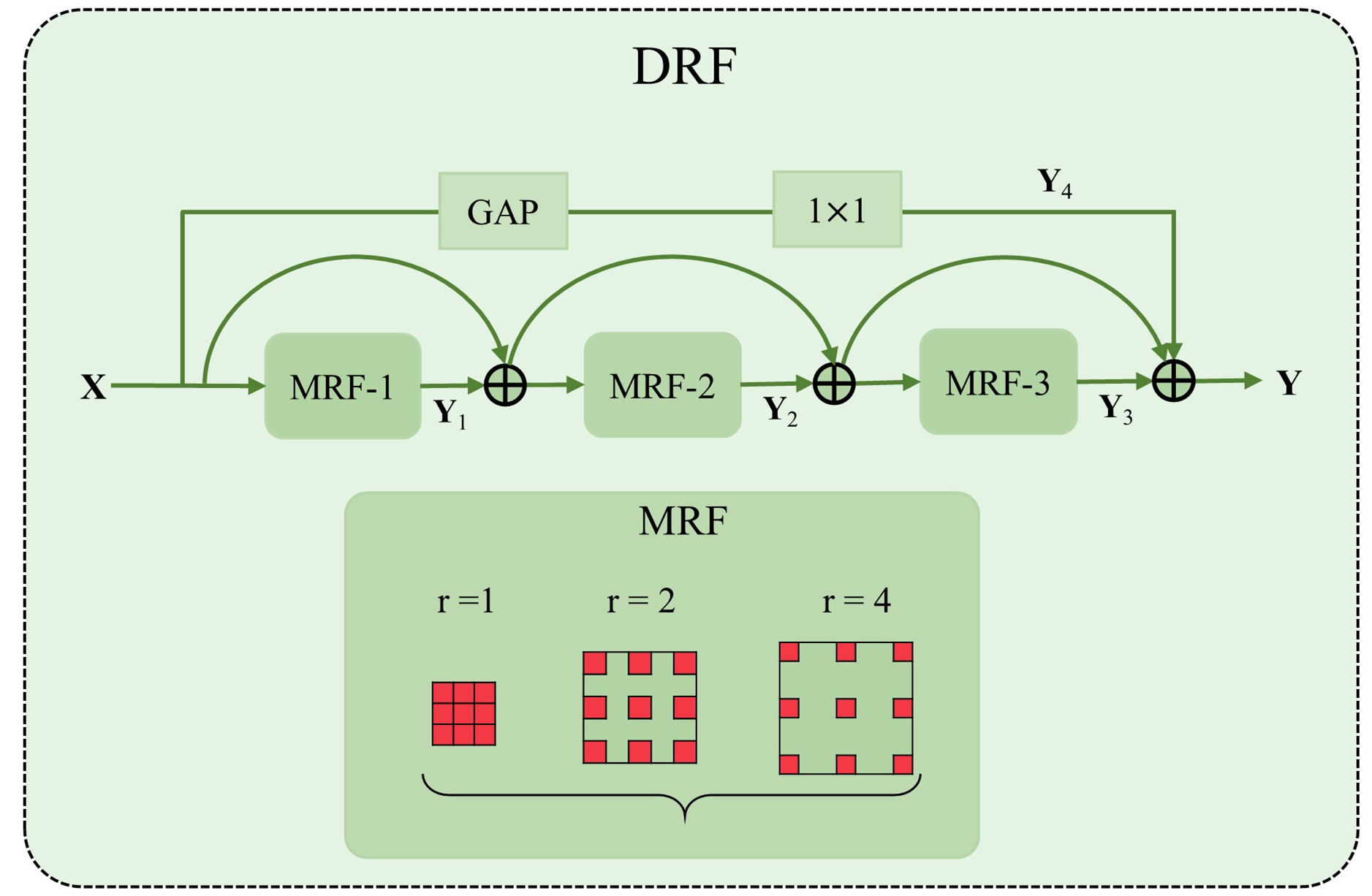}
	\caption{Illustration of the Dense Receptive Field (DRF) module. The DRF module densely connects a chain of MRF units with each consisting of three parallel 3 $\times$ 3 dilated convolutions with rates = (1, 2, 4).}
	\label{drf}
\end{figure}

As depicted in Fig.~\ref{drf}, the module employs a cascade of multi-receptive-field (MRF) units, coupled with global context features, to extract dense context information. 
A chain of MRF units imitates human visual systems, bringing rich receptive fields. And compared to the preceding  MRF unit, the subsequent MRF unit can obtain features that cover larger and denser receptive fields.
The dense skip connections among MRF units bring fast information flow and dense information interchange, which provide more flexible context modeling capability for local context information. In detail, each MRF unit takes as inputs the output features of all preceding MRF units and input features of the DRF module and aggregates them through summation. In addition, each MRF passes its output features to all subsequent MRF units.
Here, let $\mathcal{H}(x)$ and $\mathbf{X}$ term the MRF unit and the input features of the DRF module, respectively, output features of each MRF unit can be formulated as:
\begin{equation}
\label{deqn_ex1a}
\mathbf{Y}_i=\mathcal{H}_i({\mathbf{X}}+\sum_{j=1}^{i-1}\mathbf{Y}_j)(i=1, 2, 3)
\end{equation}
where $\mathbf{Y}_{i}$ denotes the output features of $i^{th}$ MRF unit.

Furthermore, the original input features $\mathbf{X}$ is performed a GAP operation in the spatial dimension, a 1 $\times$ 1 convolution layer and a bilinear upsampling function to obtain global context features $\mathbf{Y}_{4}$. Finally, the resulting multiple context information  $\mathbf{Y}_{1}$, $\mathbf{Y}_{2}$, $\mathbf{Y}_{3}$, $\mathbf{Y}_{4}$ and the original input features $\mathbf{X}$ are aggregated together through summation as the output of DRF module. Formally, output features of the DRF module can be expressed as:
\begin{equation}
\label{deqn_ex1a}
\mathbf{Y}=\mathbf{X}+\sum_{i=1}^{4}\mathbf{Y}_i
\end{equation}

In summary, the DRF module integrates local context information that covers dense receptive fields, as well as global context information. It is embedded at the end of the encoder, which further promotes the representation ability of deep features. The rich context information is beneficial to detect defects at different scales.
\subsection{Loss Function}

To obtain high-quality saliency maps with clear boundaries, we introduce hybrid loss \cite{Qin2019BASNet} to train our model. Meanwhile, we use deep supervision \cite{Hou2017deeply} on multiple side-output layers. Formally, the total loss is defined as:
\begin{equation}
\label{deqn_ex1a}
\mathcal{L}_{total}=\sum_{k=1}^{K}(\ell^{(k)}_{bce}+\ell^{(k)}_{iou}+\ell^{(k)}_{ssim})
\end{equation}
where $K$ denotes the total number of side-output layers, K = 5, including outputs from and DRF module and four decoding stages. And $\ell^{(k)}_{bce} $, $\ell^{(k)}_{iou}$ and $\ell^{(k)}_{ssim}$ denote BCE loss \cite{BoerKMR05}, IoU \cite{Rahman2016Optimizing} loss and SSIM \cite{Wang2003structural}
loss, respectively.

Given the predicted saliency map \textbf{P} and the corresponding binary ground truth \textbf{G}. BCE loss, as a pixel-wise classification loss, is defined as:
\begin{equation}
\begin{split}
    \label{deqn_ex1a}
  \ell_ {bce}  =- \frac{1}{W \times H} \sum_{(u,v)} [G(u,v) \log  (P(u,v)) + \\
  (1-G (u,v)) \log  (1-{P}(u,v))]
\end{split}
\end{equation}
where $H$ and $W$ denote the height and width of input image, and $P(u,v)\in [0,1]$ and $G (u,v) \in \{ 0,1\}$  denote predicted probability and label value of \textbf{P} and  \textbf{G} at pixel coordinates $(u,v)$, respectively.

IoU loss is calculated on the basis of the Intersection over Union (IoU) of \textbf{P} and  \textbf{G}, and it is defined as:
\begin{equation}
\label{deqn_ex1a}
 \ell_{iou}=1-\frac{\sum_{(u,v)} P(u,v) G(u,v)}{\sum_{(u,v)}[P(u,v)+G(u,v)-P(u,v) G(u,v)]}
\end{equation}

Structural similarity (SSIM) loss takes a local neighborhood of each pixel into consideration, which captures patch-level structural information. Concretely, for patch $\mathbf{R} = \left\{r_{i}: i=1, \ldots, N^{2}\right\}$  and patch $\mathbf{T}=\left\{t_{i}: i=1, \ldots, N^{2}\right\}$ cropped from \textbf{P} and \textbf{G} respectively, where N denotes the patch size, the SSIM loss between them is defined as
\begin{equation}
\label{deqn_ex1a}
 \ell_{ssim}=1-\frac{(2 \mu_{R} \mu_{T}+\varepsilon_{1})(2 \sigma_{RT}+\varepsilon_{2})}{(\mu_{R}^{2}+\mu_{T}^{2}+\varepsilon_{1})(\sigma_{R}^{2}+\sigma_{T}^{2}+\varepsilon_{2})}
\end{equation}
where $\mu_R$, $\mu_T$ and $\sigma_R$, $\sigma_T$ represent the mean and standard deviations of patch $\mathbf{R}$ and $\mathbf{T}$ respectively, $\sigma_{RT}$ represents their covariance, $\varepsilon_1 = 0.01^2$ and $\varepsilon_2= 0.03^2$.

\section{Experiments}
\subsection{Datasets}
To demonstrate the effectiveness of our method in defect detection, we evaluate different methods on three publicly available defect datasets: SD-saliency-900 dataset \cite{SONG2020Saliency}, Magnetic tile dataset \cite{Huang2018Surface}, and DAGM 2007 dataset \cite{ger2008Weakly}.

\subsubsection{SD-saliency-900} It is collected to solve the problem of automatic detection for hot-rolled steel strips, which contains three kinds of typical surface defects: patch, inclusion, and scratch. Each defect category contains 300 images with the resolution of 200 $\times$ 200, where each image is provided with the corresponding pixel-level annotation. And its complexities, such as various defect scales and types, low contrast and complex background disturbance, lead to great challenges for defect detection.

\subsubsection{Magnetic tile} It contains 1344 images with pixel-level labels, which are divided into 952 defect-free images and 392 defect images. These images with defects have different resolutions and are divided into five categories: uneven (103), fray (32), crack (57), blowhole (115), and break (85). And these defects characterized by complicated background textures, various scales and random lighting bring difficulties for automatic defect detection.

\subsubsection{DAGM 2007} It consists of 10 types of defects named from class1 to class10, which are generated by different texture models and defect models, respectively. Each category in the class1 $\sim$ class6 contains 150 defect samples, and each category in the class7 $\sim$ class10 contains 300 defect samples. Concretely, each sample with 512 $\times$ 512 resolution only contains one defect, of which the region is approximately annotated by an ellipse. And the presence of small or low-contrast defects makes this dataset challenging. 

\subsection{Data Processing}

For SD-saliency-900, like EDRNet \cite{Song2020EDRNet} and DACNet \cite{Zhou2022DACNet}, the training set is composed of 540 original images and 270 images with 20\% salt and pepper noise. The remaining 540 images are obtained by randomly selecting 180 images for each defect type in SD-saliency-900. The 270 images are generated by adding 20\% salt and pepper noise on 90 images randomly selected for each defect type in the previous 540 images. 
For DAGM 2007, we adopt the official training set and test set, which consist of 1046 images and 1054 images respectively. Before training, training set is augmented by horizontal flipping and vertical flipping. 
And for the Magnetic tile, we randomly split the training set and test set with a ratio of 1:1 for each type of defect images.
And during the training phase, each input image I is first resized to 256 $\times$ 256 resolution, then randomly cropped to 224 $\times$ 224, and fed into the network through a normalization operation $ \rm {(I - \mu ) / \sigma }$, where  $\mu$ and  $\sigma$ are equal to 0.4669 and 0.2437, respectively. The ground truth G is binarized with its maximum gray value, which is used to compute the loss with the predicted saliency map. 
And in the test phase, each input image is resized to 256 $\times$ 256 and fed into the network through the same normalization operation. Subsequently, the resulting saliency map is rescaled back to the original resolution of the input image by bilinear interpolation for evaluation.

\begin{figure*}[!t]
\centering
\includegraphics[width=5.8in,height=4.2in]{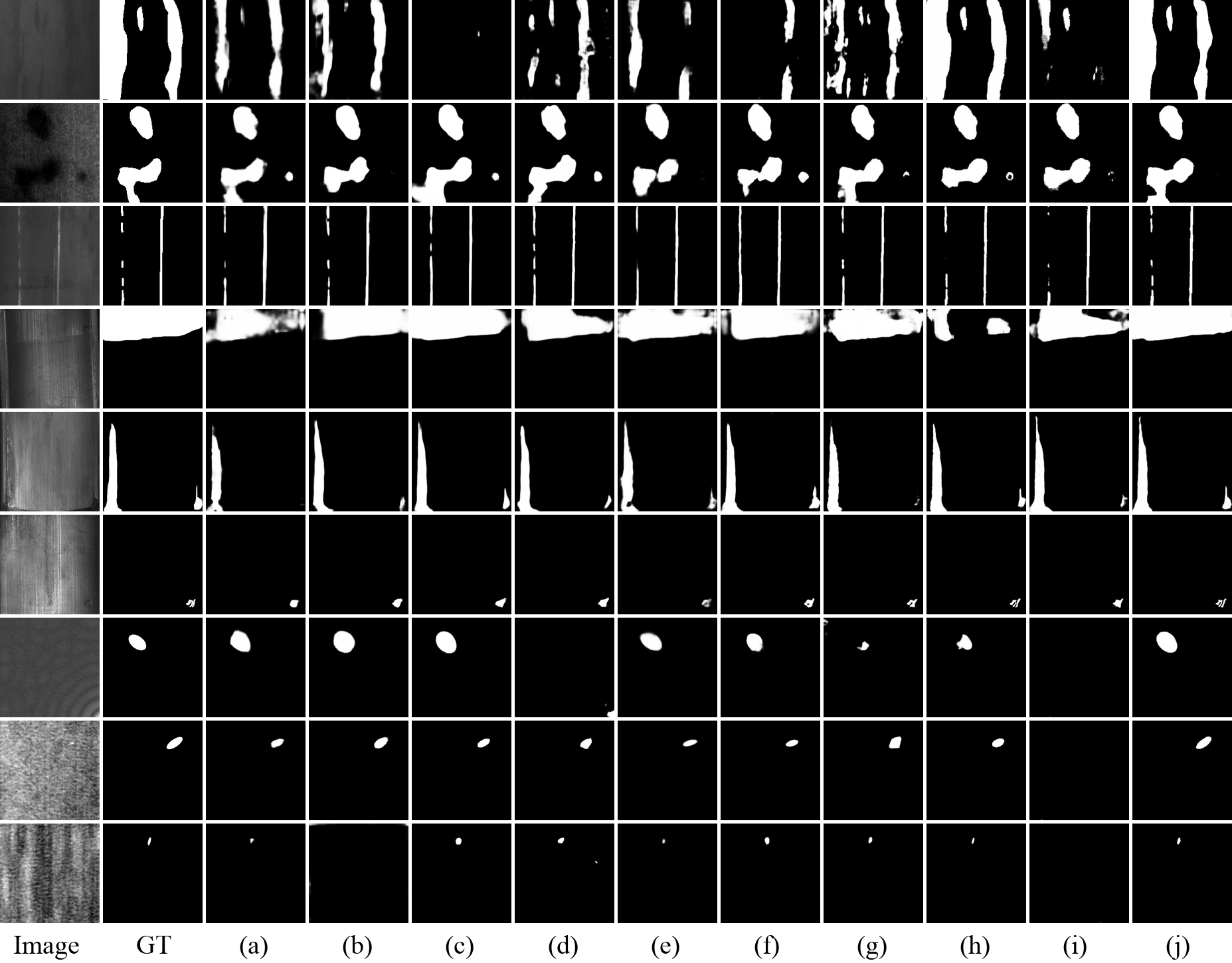}
\caption{Qualitative comparisons for saliency maps of different methods on three defect datasets: SD-saliency-900 (the 1st$\sim$3rd rows), Magnetic tile (the 4th$\sim$6th rows), DAGM 2007 (the 7th$\sim$9th rows). (a)$\sim$(j) represent LWNet, C2FNet, AttaNet, PFSNet, EDN, PGNet, CSFNet, EDRNet, DACNet and Ours, respectively.}
\label{pred}
\end{figure*}

\subsection{Implementation Details}
The network is implemented based on an open source deep learning framework: Pytorch \cite{Paszke2017Automatic}. All experiments run on the PC with an Intel eight-core CPU (3.80GHZ) and an RTX 3060 Ti GPU (with 8GB memory).
In our model, some encoder parameters are initialized with the pre-trained model of ResNet18 on ImageNet\cite{deng2009imagenet} and other parameters are initialized with the default way of Pytorch. We train our model using the Adam optimizer \cite{Kingma2018Adam} with a learning rate of 1e-3 and default values for other hyperparameters. And for SD-saliency-900, Magnetic tile, and DAGM 2007, the network is trained on their training datasets for 600 epochs, 900 epochs, and 300 epochs, respectively, with the batch size of each dataset set to 8, 5, and 8.

\subsection{Evaluation Metrics}
We adopt six common metrics for quantitative evaluation, including precision-recall (PR) curve \cite{PR}, and the F-measure curve \cite{FM}, Mean Absolute Error (MAE) \cite{MAE}, weighted F-measure ($F_{\beta}^{w}$) score \cite{WF}, structural similarity measure ($S_m$) \cite{SM}, enhanced-alignment measure ($E_{m}$) \cite{EM}. It should be noted that the smaller the MAE the better performance it indicates and vice versa for $F_{\beta}^{w}$, $S_m$, and $ E_{m}$. And PR curve and F-measure curve are two kinds of curve indicators.

PR curve \cite{PR} is frequently used to evaluate the predicted saliency map. Based on a binarized saliency map \textbf{P} and the corresponding binary ground truth \textbf{G}, we can compute a pair of precision and recall scores. Various thresholds from 0 to 1 generate a series of binarized saliency maps, on the basis of which precision-recall pairs are computed and plotted as the PR curve.

F-measure \cite{FM} is defined as the weighted harmonic mean of precision and recall. Formally, it is expressed as:
\begin{equation}
\label{deqn_ex1a}
 F_{\beta}=\frac{(1+\beta^{2}) \text {Precision} \times \text {Recall}}{\beta^{2}\times \text {Precision}+\text {Recall}}
\end{equation}
where $\beta^{2}$ is set to 0.3 as suggested in \cite{FM}. For pairs of precision and recall, we can further compute corresponding F-measure scores, on the basis of which we plotted the F-measure curve.

MAE \cite{MAE} measures the average pixel-level absolute error between \textbf{P} and \textbf{G}, which is formulated as:
\begin{equation}
\label{deqn_ex1a}
 \mathrm{MAE}=\frac{1}{W \times H} \sum_{(u,v)}|P(u,v)-G(u,v)|
\end{equation}

$F_{\beta}^{w}$ \cite{WF} is an intuitive generalization of the F-measure, which is defined as follows:
\begin{equation}
\label{deqn_ex1a}
F_{\beta}^{w}= \frac{\left(1+\beta^{2}\right)\text {Precision }^{w} \times \text { Recall }^{w}}{\beta^{2}\times \text { Precision }^{w}+\text { Recall }^{w}}
\end{equation}
where $ \beta^{2} $ is set to 1 as suggested in \cite{WF}.

$S_m$ \cite{SM} simultaneously evaluates object-aware and region-aware structural similarities between \textbf{P} and \textbf{G}, denoted as $S_{o}$ and $S_{r}$, respectively. It is formulated as:
\begin{equation}
\label{deqn_ex1a}
 S_m=\lambda \times S_{o}+(1-\lambda) \times S_{r}
\end{equation}
where $\lambda$ is set to 0.5 as suggested in \cite{SM}.

$E_{m}$ \cite{EM} simultaneously takes global statistics and local pixel matching into consideration and is defined as:
\begin{equation}
\label{deqn_ex1a}
 E_{m}=\frac{1}{W \times H} \sum_{(u,v)} \phi(u,v)
\end{equation}
where $\phi$ denotes the enhanced alignment matrix, reflecting the correlation between \textbf{P} and \textbf{G} subtracted each global means.

\subsection{Experimental Results}
We compare our method with other state-of-the-art saliency detection models or scene segmentation models, containing C2FNet \cite{Sun2021C2FNet}, PGNet \cite{Xie2022PGNet}, DACNet \cite{Zhou2022DACNet}, EDRNet \cite{Song2020EDRNet}, LWNet \cite{Huang2020Compact}, AttaNet \cite{Song2021AttaNet}, PFSNet \cite{Ma2021Pyramidal}, EDN \cite{wu2022edn}, and CSFNet \cite{gao2020sod100k}. Note that, \cite{Song2020EDRNet,Huang2020Compact,Zhou2022DACNet} are specifically designed for surface defect detection tasks.

\subsubsection{Results on SD-saliency-900} The top three rows in Fig.~\ref{pred} display partial detection results of different methods on SD-saliency-900 disturbed by salt-and-pepper noise. The addition of interference noise not only brings a lot of background noise but also weakens the contrast between the defects and the background, which makes detection more challenging. 
And for these complex defects, most methods merely detect partial defect regions, such as the 1st row. Some methods even falsely detect background disturbances as defects, such as the 2nd row. In contrast, our method obtains more accurate and complete saliency maps.

\begin{figure*}[!t]
\centering
\includegraphics[width=5.9in,height=3in]{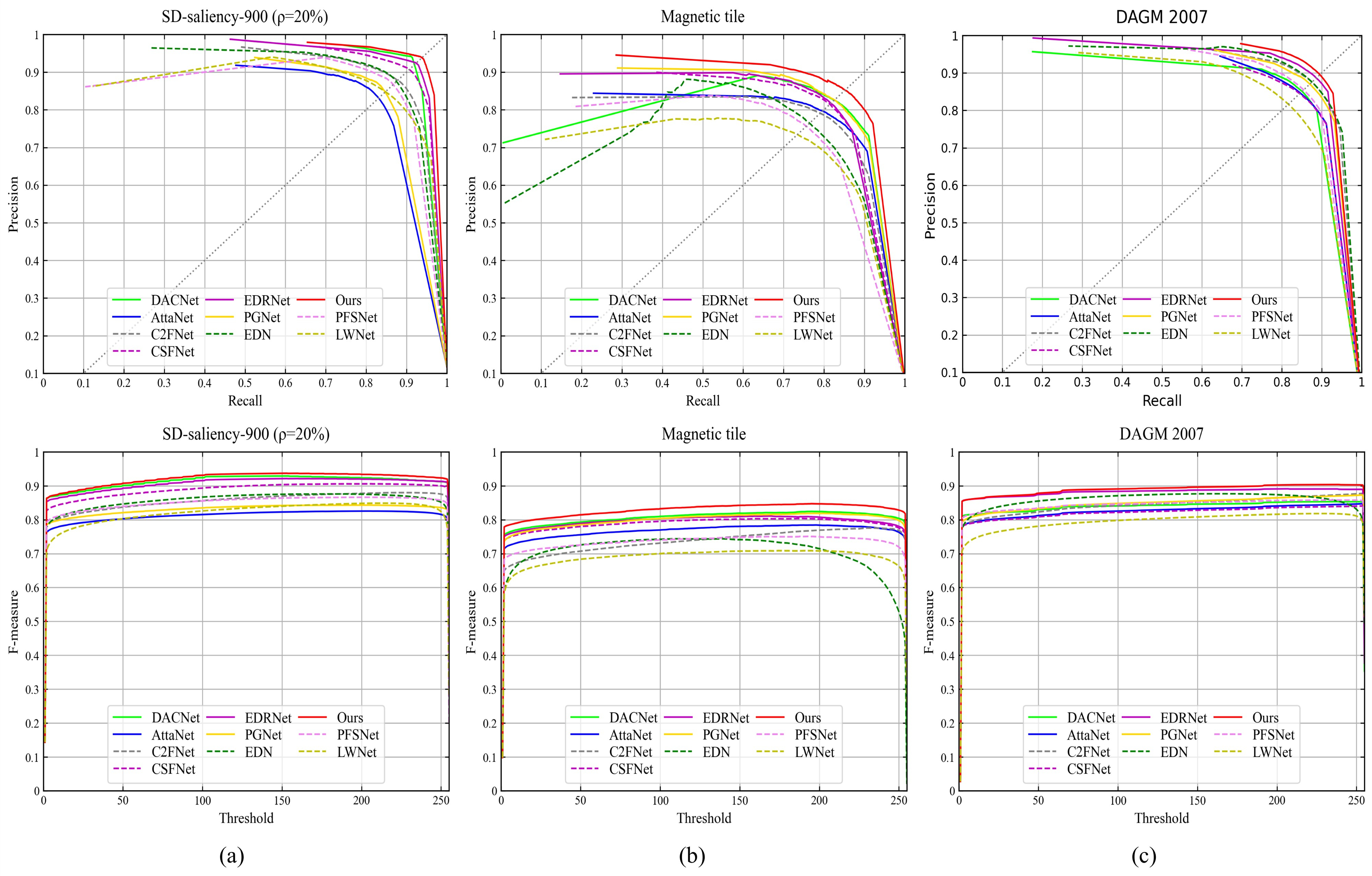}
\caption{The PR curves and F-measure curves of different methods on three defect datasets. (a), (b) and (c) represent SD-saliency-900 ($\rho$=20\%), Magnetic tile and DAGM 2007, respectively.}
\label{curves}
\end{figure*}

In addition, as listed in quantitative comparisons in Table~\ref{tab1}, our method yields impressive results compared to other methods. Notably, even compared to \cite{Zhou2022DACNet} and \cite{Song2020EDRNet}, our method also brings significant improvements. Especially, compared to the \cite{Song2020EDRNet}, our model obtains a gain of 1.66 \% in terms of $F_{\beta}^{w}$. And our model also obtains gains of 11.11\%, 1.31\%, and 1.17\% on MAE, $F_{\beta}^{w}$, and $S_m$ respectively, compared with the \cite{Zhou2022DACNet}. Meanwhile, it is also observed from Fig.~\ref{curves} (a) that the PR curve and F-measure curve by our method almost cover other curves, indicating our method obtains higher precision and higher F-measure score under most thresholds.

\begin{table*}[!t]
\renewcommand\arraystretch{1.1}
\centering
\caption{Quantitative comparisons on SD-saliency-900 ($\rho$=20\%), Magnetic tile, and DAGM 2007 in terms of MAE, $F_{\beta}^{w}$, $S_m$ and $E_{m}$. $"\uparrow" / "\downarrow"$  indicates that larger or smaller is better. The best results on each dataset are highlighted in bold.}
\label{tab1}
\setlength{\tabcolsep}{1.8mm}{
\begin{tabular}{r|r|cccc|cccc|cccc}		
\hline
\hline
\multirow{2}*{Methods} &\multirow{2}*{Backbone}&\multicolumn{4}{c|}{\tabincell{c}{SD-saliency-900\\($\rho=20\%$)}} & \multicolumn{4}{c|}{Magnetic tile} & \multicolumn{4}{c}{DAGM 2007}\\
\cline{3-14}
~&~&$\text{MAE}\downarrow$ &$F_{\beta}^{w}\uparrow$&$S_m\uparrow$&$E_{m}\uparrow$ & $\text{MAE}\downarrow$ &$F_{\beta}^{w}\uparrow$&$S_m\uparrow$&$E_{m}\uparrow$ & $\text{MAE}\downarrow$ &$F_{\beta}^{w}\uparrow$&$S_m\uparrow$&$E_{m}\uparrow$\\
\hline
LWNet\cite{Huang2020Compact}&\makecell[c]{---} &0.0260& 0.8166&0.8688 &0.9253&0.0232&0.6736&0.8010&0.7753&0.0071& 0.7629&0.8644 &0.9090\\
C2FNet \cite{Sun2021C2FNet}&Res2Net50& 0.0210& 0.8515&0.8897 &0.9433&0.0205&0.7320&0.8376&0.8361&0.0052& 0.8345&0.9039 &0.9357\\
AttaNet\cite{Song2021AttaNet} & ResNet18&0.0256& 0.8011&0.8537 &0.9227&0.0179&0.7613&0.8461&0.8858&0.0077 & 0.8050 & 0.8827 & 0.9360\\
PFSNet\cite{Ma2021Pyramidal} &ResNet50& 0.0230& 0.8413&0.8803 &0.9402&0.0211& 0.7203&0.8287&0.8822&0.0058 & 0.8193 & 0.8842 & 0.9470\\
EDN\cite{wu2022edn}  &ResNet50&0.0201& 0.8466&0.8844 &0.9461&0.0195& 0.6832&0.8133&0.7786&0.0050 & 0.8318 & 0.9042 & 0.9358\\
PGNet \cite{Xie2022PGNet}& \makecell[c]{---}&0.0238& 0.8190&0.8628 &0.9279&0.0174&0.7894&0.8632&0.8965&0.0056 & 0.8345 & 0.8999 & 0.9443\\
CSFNet\cite{gao2020sod100k}  &Res2Net50 &0.0170& 0.8863&0.9106 &0.9559&0.0181&0.7792&0.8625&0.8977&0.0086& 0.8033&0.8851 &0.9325\\
EDRNet\cite{Song2020EDRNet}  & ResNet34& 0.0146& 0.9056&0.9244 &0.9638&0.0204&0.7828&0.8601&0.9038&0.0046 & 0.8612 & 0.9097 & 0.9626\\
DACNet\cite{Zhou2022DACNet} &ResNet34&0.0144& 0.9087&0.9238 &0.9658&0.0194&0.7993&0.8676&0.9135&0.0065& 0.8227&0.8914 &0.9519\\
\hline
Ours  &ResNet18 & \textbf{0.0128}& \textbf{0.9206}&\textbf{0.9346}&\textbf{0.9682}& \textbf{0.0172}&\textbf{0.8205}& \textbf{0.8798}& \textbf{0.9202}& \textbf{0.0042}& \textbf{0.8752}&\textbf{0.9203} &\textbf{0.9645}\\
\hline
\hline
\end{tabular}}
\end{table*}

\subsubsection{Results on Magnetic tile} The 4th $\sim$ 6th rows in Fig.~\ref{pred} present partial detection results of different methods on Magnetic tile. 
As shown in Fig.~\ref{pred}, some defects appear highly similar to the background (the 4th and 5th rows) or small (the 6th row), making it difficult to determine their locations. For these challenging defects, those existing methods may suffer from several problems, such as detecting only part of the defect regions, missing the weak or tiny defects, mistakenly recognizing the background regions as defects, and so on.
However, our method obtains detection results similar to ground truths.
And as shown in Fig.~\ref{curves} (b), the PR and F-measure curves by our method are both able to completely cover those of other methods, which indicates our method has an obvious advantage over other methods on precision and F-measure. Moreover, as listed in Table~\ref{tab1}, our model also obtains performance gains of 2.65\% and 1.41\% over \cite{Zhou2022DACNet} on $F_{\beta}^{w}$ and $S_m$ respectively, the MAE is reduced from 0.0194 to 0.0172, obtaining a sharp decrease of 11.34\%.

\subsubsection{Results on DAGM 2007} The 7th $\sim$ 9th rows in Fig.~\ref {pred} display partial detection results of different methods on DAGM 2007. The selected defects are characterized by low contrast (the 7th and 8th rows) or small size (the 9th row). From detection results, we can find some methods, such as \cite{Sun2021C2FNet}, \cite{Zhou2022DACNet} and \cite{Ma2021Pyramidal}, may miss these defects directly.
In contrast, our method obtains the best prediction results for these challenging defects. Meanwhile, as reported in Table~\ref{tab1}, the performance of our method ranks first in the comparison with other methods. For example, compared with the \cite{Song2020EDRNet}, our model improves by 1.63\% and 1.17\% in terms of $F_{\beta}^{w}$ and $S_m$ respectively. Similarly, the PR curve and F-measure curve by our method (the red curve) outperform those of other methods in most cases, as shown in Fig.~\ref{curves} (c).

\subsubsection{Running Efficiency} we further evaluate the running efficiency of various methods on the SD-saliency-900 dataset using model size, params, floating-point operation numbers (FLOPs), and inference speed (FPS).
The model size, FLOPs and FPS reflect the memory space, computation overhead and detection speed of the model, respectively.
Concretely, we evaluate the running efficiency of different models with the input size of 256 $\times$ 256 and the batch size of  1. 
The experimental results are reported in Table~\ref{tab2}.
Although \cite{Zhou2022DACNet} obtains significant performance improvement compared to \cite{Song2020EDRNet} on SD-saliency-900, it brings higher complexity. In contrast, our method only increases fewer parameters but achieves more significant performance gains compared with \cite{Song2020EDRNet}.
It is easily found that our method achieves a considerable detection speed when the batch size is 1. It takes about 0.015s to process an image with 256 $\times$ 256 resolution.
And as listed in Table~\ref{tab2}, an obvious drawback of our model is that it has more parameters and computational cost than \cite{Sun2021C2FNet}, \cite{Song2020EDRNet}, \cite{Ma2021Pyramidal}, \cite{wu2022edn} and \cite{gao2020sod100k}.
In future work, we aim to reduce the complexity of our model and make it lightweight.

\begin{table*}[!t]
\renewcommand\arraystretch{1.2}
\centering
\caption{Comparisons of model size, params, FLOPS and speed of different methods on SD-saliency-900.}
\label{tab2}
\setlength{\tabcolsep}{1.2mm}{
\begin{tabular}{l|cccccccccc}		
\hline
\hline
Methods&\tabincell{c}{LWNet\\\cite{Huang2020Compact}} & 
\tabincell{c}{C2FNet \\ \cite{Sun2021C2FNet}}& 
\tabincell{c}{AttaNet\\\cite{Song2021AttaNet}}&
\tabincell{c}{PFSNet\\\cite{Ma2021Pyramidal}} &
\tabincell{c}{EDN\\ \cite{wu2022edn}} &
\tabincell{c}{PGNet \\\cite{Xie2022PGNet}}& 
\tabincell{c}{CSFNet\\\cite{gao2020sod100k}}& 
\tabincell{c}{EDRNet\\\cite{Song2020EDRNet}} &
\tabincell{c}{DACNet\\\cite{Zhou2022DACNet}} & 
\tabincell{c}{Ours}\\
\hline
Model size (MB)&8.64&108&50.8&119&164&279&139&150&375&160\\
\hline
Params (M)&2.21&26.36&12.78&31.18&42.85&72.62&29.13&39.31&98.39&41.94\\
\hline
FLOPs (G) &0.46&6.92&2.95&24.02&9.05&18.37&7.87&42.14&142.71&60.19\\
\hline
Speed (FPS) &71&38&126&20&34&33&48&35&39&66\\
\hline
\hline
\end{tabular}}
\end{table*}

\begin{table*}[!t]
\renewcommand\arraystretch{1.2}
\centering
\caption{Architecture ablation analysis on SD-saliency-900 ($\rho$=20\%), Magnetic tile, and DAGM 2007. Ours denotes the full version of our proposed method.}
\label{tab3}
\setlength{\tabcolsep}{1.5mm}{
\begin{tabular}{c|r|cccc|cccc|cccc}		
\hline
\hline

\multirow{2}*{Module}&\multirow{2}*{Settings} &\multicolumn{4}{c|}{\tabincell{c}{SD-saliency-900\\($\rho=20\%$)}} & \multicolumn{4}{c|}{Magnetic tile} & \multicolumn{4}{c}{DAGM 2007}\\
\cline{3-14}
~&~&$\text{MAE}\downarrow$ &$F_{\beta}^{w}\uparrow$&$S_m\uparrow$ & $E_{m}\uparrow$ &$\text{MAE}\downarrow$ &$F_{\beta}^{w}\uparrow$&$S_m\uparrow$& $E_{m}\uparrow$ &$\text{MAE}\downarrow$ &$F_{\beta}^{w}\uparrow$&$S_m\uparrow$&$E_{m}\uparrow$ \\
\hline
 &Baseline &0.0159 & 0.8981 & 0.9204&0.9605&0.0193&0.7953&0.8675&0.8971& 0.0105 & 0.8151 & 0.8924&0.9283\\
\hline
\multirow{6}*{JAFF}&w/o JAFF & 0.0150 & 0.9067 & 0.9250&0.9628&0.0184 & 0.8034 & 0.8699 &0.9029&0.0051&0.8542&0.9115&0.9542\\
~&w/o SAB &  0.0138 & 0.9148 & 0.9309&0.9660& 0.0170 & 0.8134 & 0.8754&0.9157&0.0048&0.8672&0.9163&0.9632\\
~&w/o CAB & 0.0137 & 0.9127 & 0.9304&0.9649&\textbf{0.0165}&0.8105&0,8719&0.9158&0.0045&0.8685&0.9175&0.9613 \\
\cline{2-14}
~&w SE \cite{hu2018squeeze}&0.0140 & 0.9136 & 0.9295&0.9658 &0.0174&0.8117&0.8720&0.9148&0.0049 & 0.8635 & 0.9146&0.9615\\
~&w CBAM \cite{woo2018cbam}&0.0140&0.9134&0.9297&0.9666&0.0182 & 0.8136 & 0.8742&0.9187&0.0045&0.8693&0.9174&0.9634\\
~&w BAM \cite{park2018bam}&0.0135 & 0.9167 & 0.9309&0.9670&0.0178 & 0.8126 & 0.8739&0.9165&0.0047&0.8699&0.9162&0.9628\\
\hline
\multirow{3}*{DRF}&w/o DRF &0.0138 & 0.9125 & 0.9289& 0.9629&0.0182&0.8061&0.8749&0.9100&0.0060&0.8411&0.9054&0.9563\\
~&w PPM (1,2,3,6) \cite{Zhao2017Pyramid}  &0.0138 & 0.9131 & 0.9314&0.9657&0.0177&0.8112&0.8755&0.9165&0.0051&0.8573&0.9134& 0.9592 \\
~&w ASPP (1,2,4) \cite{Chen2017Rethinking}  &0.0136 & 0.9139 & 0.9303&0.9659&0.0177&0.8098&0.8735&0.9162&0.0054&0.8558&0.9112& 0.9587\\
\hline
~&Ours & \textbf{0.0128}& \textbf{0.9206}&\textbf{0.9346} &\textbf{0.9682}
&0.0172&\textbf{0.8205}& \textbf{0.8798}&\textbf{0.9202}&\textbf{0.0042}& \textbf{0.8752}&\textbf{0.9203}&\textbf{0.9645 }
\\
\hline
\hline
\end{tabular}}
\end{table*}

\begin{table*}[!t]
\renewcommand\arraystretch{1.2}
\centering
\caption{Loss ablation analysis on SD-saliency-900 ($\rho$=20\%), Magnetic tile, and DAGM 2007.}
\label{tab4}
\setlength{\tabcolsep}{1.5mm}{
\begin{tabular}{r|cccc|cccc|cccc}		
\hline
\hline
\multirow{2}*{Settings} &\multicolumn{4}{c|}{\tabincell{c}{SD-saliency-900\\($\rho=20\%$)}} & \multicolumn{4}{c|}{Magnetic tile} & \multicolumn{4}{c}{DAGM 2007}\\
\cline{2-13}
~&$\text{MAE}\downarrow$ &$F_{\beta}^{w}\uparrow$&$S_m\uparrow$ & $E_{m}\uparrow$ &$\text{MAE}\downarrow$ &$F_{\beta}^{w}\uparrow$&$S_m\uparrow$& $E_{m}\uparrow$ &$\text{MAE}\downarrow$ &$F_{\beta}^{w}\uparrow$&$S_m\uparrow$&$E_{m}\uparrow$\\
\hline
$\ell_{bce}$ &0.0142 & 0.9078 & 0.9275&0.9628&0.0179 & 0.7777 & 0.8681&0.8396&0.0073&0.8407&0.9084&0.9350 \\
$\ell_{iou}$&0.0145 & 0.9106 & 0.9256&0.9648&\textbf{0.0161}&0.8036&0.8677&0.9152&0.0052&0.8558&0.9102&0.9597 \\
$\ell_{ssim}$ & 0.0147 & 0.9105 & 0.9267&0.9637& 0.0182 & 0.7819 & 0.8635&0.8817&0.0063&0.8330&0.9001&0.9460 \\
$\ell_{bce} + \ell_{iou}$ & 0.0136 & 0.9150 & 0.9301&0.9658&0.0177&0.8120&0.8737&0.9201& 0.0050 & 0.8654 & 0.9139&0.9630\\
$\ell_{bce} + \ell_{ssim}$ &0.0140 & 0.9113 & 0.9299&0.9626&0.0175&0.7925&0.8738&0.8749&0.0048&0.8565&0.9169&0.9500\\
w/o DP & 0.0149 & 0.9037 & 0.9214 &0.9645&0.0196&0.7940&0.8645 &0.9096&0.0061&0.8433&0.9054&0.9541\\
\hline
Ours &  \textbf{0.0128}& \textbf{0.9206}&\textbf{0.9346}&\textbf{0.9682}&0.0172&\textbf{0.8205}& \textbf{0.8798} &\textbf{0.9202}&\textbf{0.0042}& \textbf{0.8752}&\textbf{0.9203}&\textbf{0.9645}\\
\hline
\hline
\end{tabular}}
\end{table*}

\begin{figure*}[!t]
\centering
\includegraphics[width=6.3in,height=2.8in]{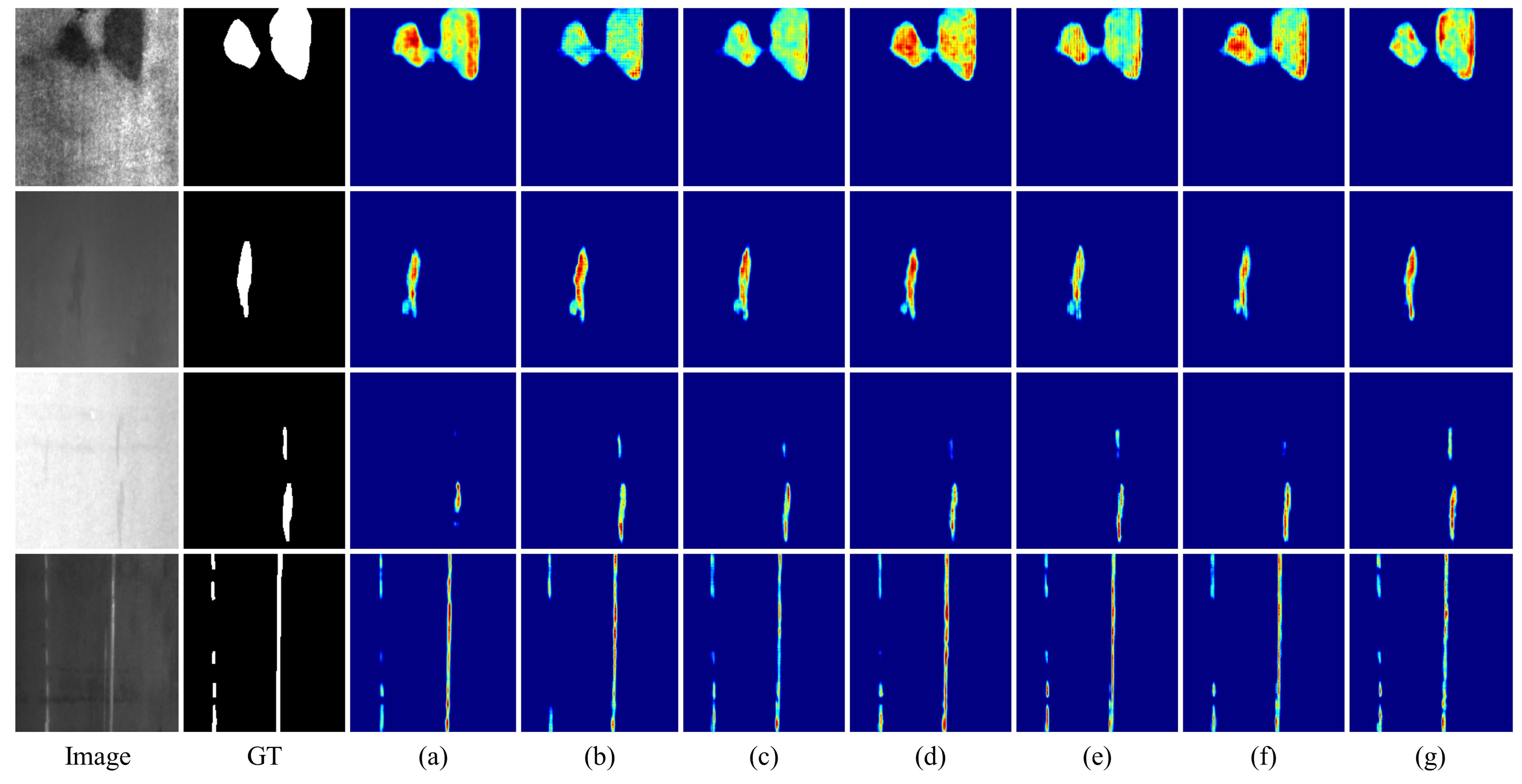}
\caption{Visual comparisons for feature maps generated by the last decoding stage $D_4$ of different variants. (a) $\sim$ (g) represent the variants w/o JAFF, w/o SAB, w/o CAB, w SE, w CBAM, w BAM and Ours, respectively. The dark blue regions represent background, while the bright regions represent predicted defect regions.}
\label{ablation1}
\end{figure*}

\begin{figure}[!t]
\centering
\includegraphics[width=3.3in,height=2.3in]{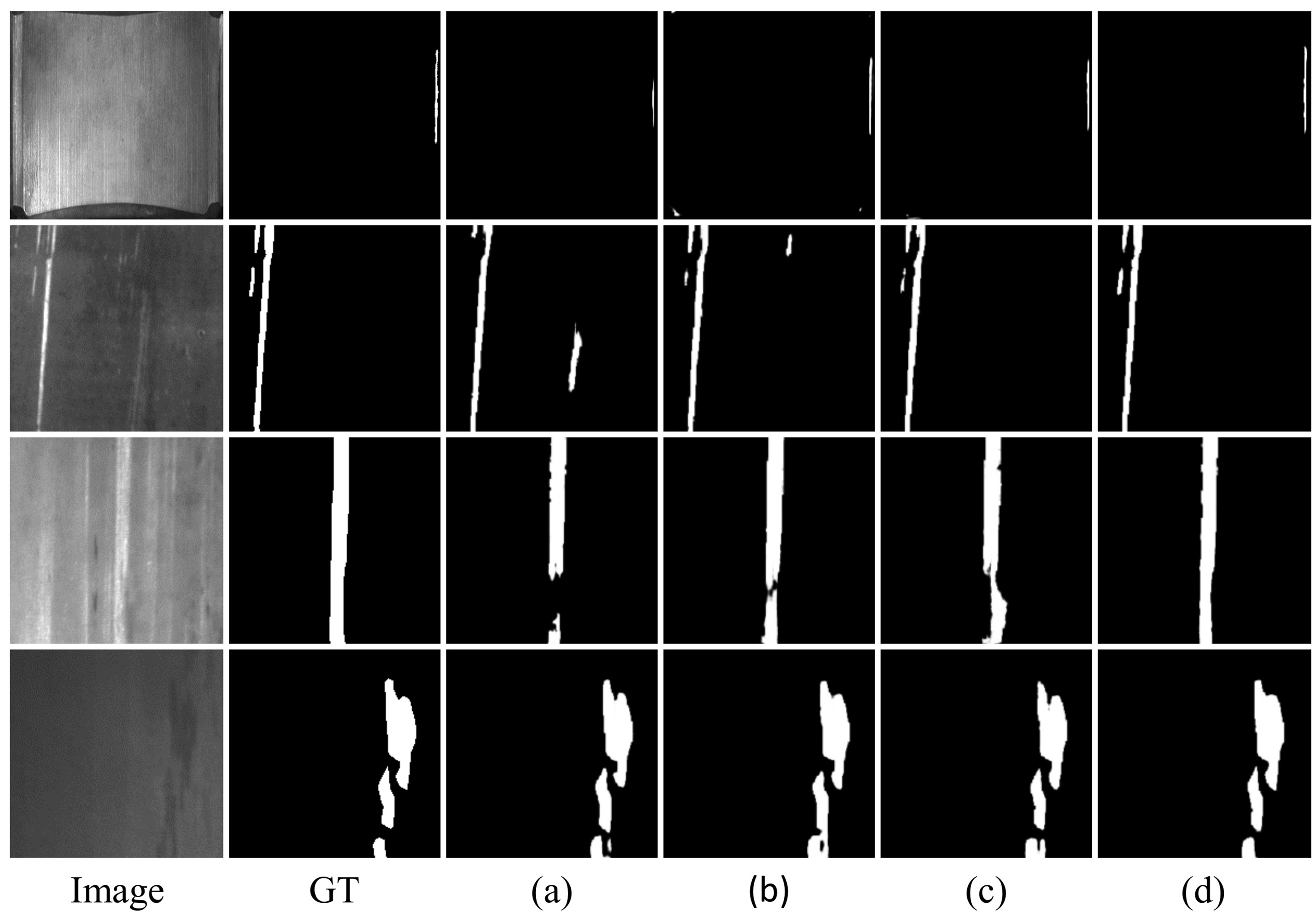}
\caption{Visual comparisons for saliency maps generated by different variants. (a) $\sim$ (d) represent the variants w/o DRF, w PPM, w ASPP and Ours, respectively.}
\label{ablation2}
\end{figure}

\subsection{Ablation Analysis}
We conduct ablation experiments for network architecture and loss, respectively, based on the above three defect datasets. 

\subsubsection{Architecture Ablation} 
To verify the necessity of each module in the network, a series of experiments are performed on networks with different settings. 
Specifically, we use the model without JAFF module and DRF module as the baseline, which directly concatenates the upsampled high-level features and the corresponding low-level features during feature fusion and replaces the DRF module with two res-basic blocks
In addition, we design the following variants based on our model: (1) model with DRF module being replaced with two res-basic blocks (w/o DRF); (2) model with JAFF module being replaced with direct concatenation operation (w/o JAFF); (3) model without channel attention branch (w/o CAB); (4) model without spatial attention branch (w/o SAB). In addition, we compare the dual attention module in JAFF with SE module \cite{hu2018squeeze}, CBAM \cite{woo2018cbam} and BAM \cite{park2018bam} to demonstrate its advantage, denoted as variants w SE, w CBAM and w BAM, respectively. We also replace DRF module with PPM \cite{Zhao2017Pyramid} and ASPP module \cite{Chen2017Rethinking} to verify its superiority, represented as variants w PPM and w ASPP, respectively.  And we evaluate these variants on MAE, $F_{\beta}^{w}$, $S_m$ and $E_m$. The detailed quantitative evaluation results are reported in Table~\ref{tab3}.

In general, our model achieves significant improvements over the baseline (the 1st row), achieving average improvements of 30.13\%, 4.35\%, 2.03\% and 2.42\%  in terms of MAE, $F_{\beta}^{w}$, $S_m$ and $E_m$, respectively.
Specifically, compared with the w/o JAFF, Ours obtains average improvements of 2.04\%, 1.05\% and 1.19\% in terms of $F_{\beta}^{w}$, $S_m$ and $E_m$, respectively. The JAFF can highlight low-level defect features and reduce background interference during feature fusion, thus generating defect-focused feature maps, as shown Fig. \ref{ablation1} (a) and (g). Ours also outperforms the variants w SE, w CBAM and w BAM on $F_{\beta}^{w}$ and $S_m$, indicating the superiority of dual attention module. Fig. \ref{ablation1} (d) $\sim$ (f) also display the feature maps generated by the last decoding stage $D_4$ of these variants. The 1st and 2nd rows indicate that Ours can significantly reduce background noise when background of defects is complicated. The 3rd and 4th rows indicate that Ours can focus on more defect details during feature fusion when defects appear small or low-contrast.

Besides, compared to the w/o DRF, Ours improves by 2.24\%, 0.94\% and 0.84\% on average in terms of $F_{\beta}^{w}$, $S_m$ and $E_m$, respectively. Ours is also superior to the variants w PPM and w ASPP on $F_{\beta}^{w}$. The main reason is that compared with PPM and ASPP, the DRF brings richer receptive fields for the model. The rich context information is beneficial to detect defects at different scales. For example, we can see from Fig.\ref{ablation2} that the DRF can accurately detect some small defects (the 1st and 2nd rows) and deal with some large-scale defects (the 3rd and 4th rows).

\subsubsection{Loss Ablation} 
We also conduct the following experiments to demonstrate the effectiveness of deep supervision and hybrid loss during the training process. 
To be specific, we compare the performance of models trained with different loss functions, as well as the model trained without deep supervision (denoted as w/o DP).
The experimental results are reported in Table~\ref{tab4}. 
It is observed that compared with the model trained with the BCE loss, Ours obtains average improvements of 18.75 \%, 3.67 \%,1.14 \% and 4.44\% on MAE, $F_{\beta}^{w}$, $S_m$ and $E_m$, respectively. 
Meantime, compared to Ours, the performance of w/o DP drops a lot. 
The experimental results suggest that hybrid loss and deep supervision play an important role during the training process.

\begin{figure}[!t]
\centering
\includegraphics[width=3.4in,height=1.8in]{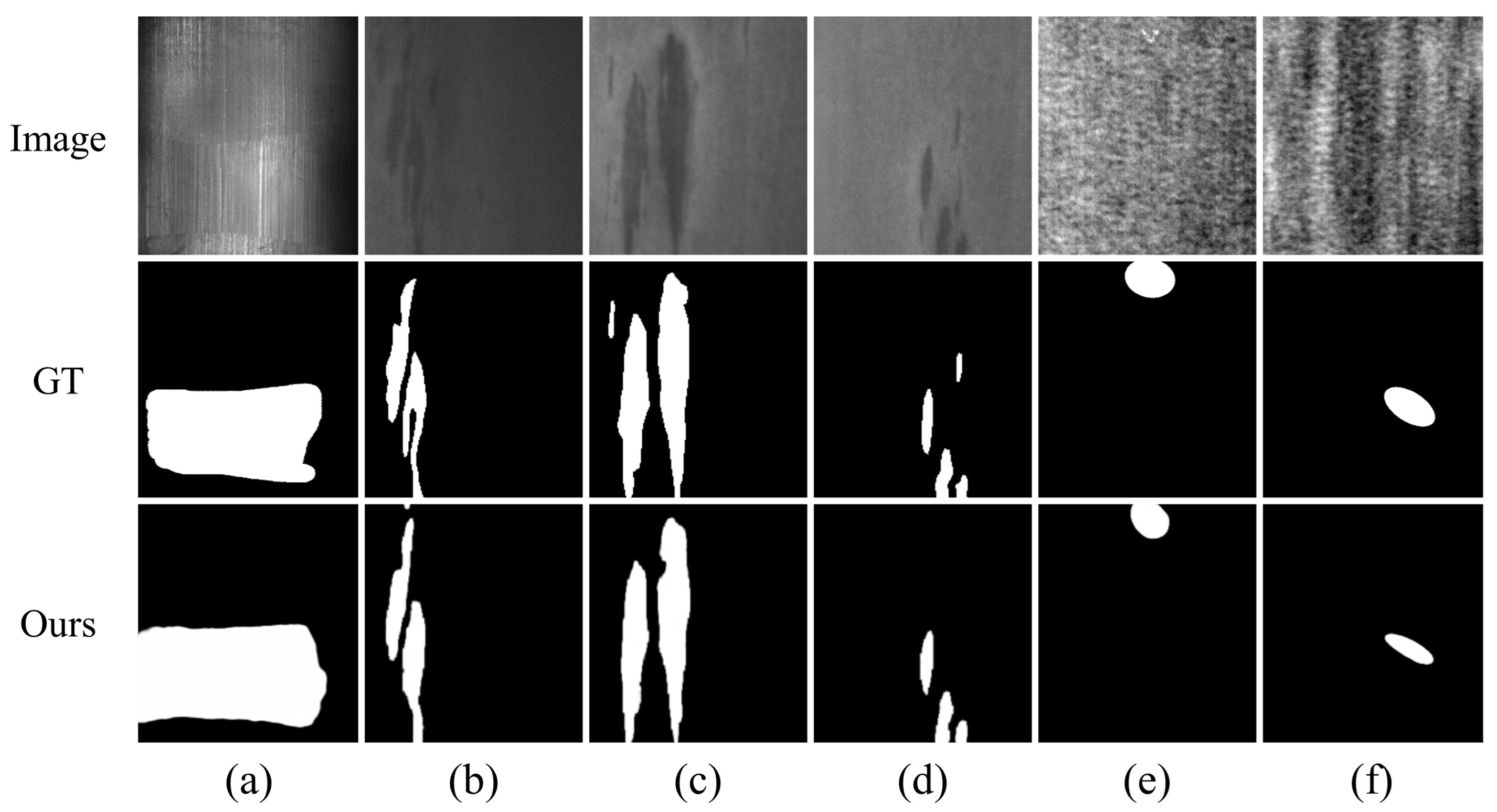}
\caption{Some representative failure cases of the proposed method.}
\label{failure}
\end{figure}

\subsection{Failure Case Analysis}
\indent Although the proposed method achieves accurate segmentation results for most scenarios, it struggles to address some very challenging defects. For example, the proposed method may detect the background regions around some defects as defects, as shown in Fig. \ref{failure} (a) and (b). The proposed method may miss some tiny defects, as shown in Fig. \ref{failure} (c) and (d). The proposed method may fail to segment out the complete defect regions, as shown in Fig. \ref{failure} (e) and (f).
The main reason for these failure cases is the lack of defect samples. 
When the number of some types of defects is few, this can lead to large differences in the distribution of the training and test datasets. This difference can affect the generalization ability of the model, bringing some failure cases.
These failure cases also suggest that there is still much room for improvement in our model.

\section{Conclusion}
In this paper, we have proposed a joint attention-guided feature fusion network for saliency detection of surface defects. Concretely, we present a novel JAFF module to guide the fusion of high-level and low-level features. 
JAFF learns a channel-spatial attention map to highlight the detailed features of weak defects and suppress background interference during feature fusion.
Besides, we present a novel DRF module that can capture context information with a large and dense range of receptive fields, which can deal with defects of various scales.
The experiments on three publicly available defect datasets indicate that JAFFNet has achieved promising performance, which proves the effectiveness of our method in defect detection. Meanwhile, it can run at a real-time speed of 66 FPS on a single 3060 Ti GPU. 
In our future research, we aim to reduce the complexity of the model through network compression techniques and make it practical in resource-constrained industrial applications.
\bibliographystyle{ieeetr}
\bibliography{ref.bib}
\vspace{-5 mm}
\begin{IEEEbiographynophoto}
{Xiaoheng Jiang} received the B.S., M.S., and Ph.D. degrees in electronic information engineering from Tianjin University, Tianjin, China, in 2010, 2013, and 2017, respectively.
\par
\hspace{-1em} He is currently an Associate Professor with the School of Computer and Artificial Intelligence, Zhengzhou University, Zhengzhou, China. His research interests include computer vision and deep learning.
\end{IEEEbiographynophoto}
\vspace{-10 mm}
\begin{IEEEbiographynophoto}
{Feng Yan} received the B.S. degree in information engineering from East China Jiao Tong University, Nanchang, China, in 2020. He is currently pursuing the M.S. degree with the School of Computer and Artificial Intelligence, Zhengzhou University, Zhengzhou, China.
\par
\hspace{-1em} His current research interests include salient object detection, semantic segmentation, deep learning, and computer vision.
\end{IEEEbiographynophoto}
\vspace{-10 mm}
\begin{IEEEbiographynophoto}
{Yang Lu} received the B.S. and Ph.D. degrees in communication engineering from Jilin University, Changchun, China, in 2014 and 2019, respectively. 
\par
\hspace{-1em} She is currently a Lecturer with the School of Computer and Artificial Intelligence, Zhengzhou University, Zhengzhou, China. Her research interests include computer vision, image processing and artificial intelligence.
\end{IEEEbiographynophoto}
\vspace{-10 mm}
\begin{IEEEbiographynophoto}
{Ke Wang} received the B.S. degree in software engineering and the M.S. and Ph.D. degrees in computer science from Beijing Institute of Technology, Beijing, China, in 2007, 2009, and 2017, respectively. 
\par
\hspace{-1em} He is currently a Lecturer with the School of Computer and Artificial Intelligence, Zhengzhou University, Zhengzhou, China. His research interests include computational intelligence, machine learning theories, algorithms and applications.
\end{IEEEbiographynophoto}
\vspace{-10 mm}
\begin{IEEEbiographynophoto}
{Shuai Guo} received the B.S. degree from Dalian University of technology in 2006, Ph.D. degree in mechanical electronics from University of Chinese Academy of Sciences, Beijing, China, in 2013.    
\par
\hspace{-1em} He is currently a Lecturer with the School of Computer and Artificial Intelligence, Zhengzhou University, Zhengzhou, China. His research interests include simultaneous localization and mapping (SLAM), deep learning, 3D point cloud processing.
\end{IEEEbiographynophoto}
\vspace{-10 mm}
\begin{IEEEbiographynophoto}
{Tianzhu Zhang} received the bachelor’s degree in communications and information technology from the Beijing Institute of Technology, Beijing, China, in 2006, and the Ph.D. degree in pattern recognition and intelligent systems from the Institute of Automation, Chinese Academy of Sciences (CASIA), Beijing, in 2011.  
\par
\hspace{-1em} He is currently a Professor with the Department of Automation, School of Information Science and Technology, University of Science and Technology of China. His current research interests include pattern recognition, computer vision, multimedia computing, and machine learning.
\end{IEEEbiographynophoto}
\vspace{-10 mm}
\begin{IEEEbiographynophoto}
{Yanwei Pang} (Senior Member, IEEE) received the Ph.D. degree in electronic engineering from the University of Science and Technology of China, Hefei, China, in 2004. 
\par
\hspace{-1em} He is currently a Professor with the School of Electrical and Information Engineering, Tianjin University, Tianjin, China. He has authored over 120 scientific articles. His current research interests include object detection and recognition, vision in bad weather, and computer vision.
\end{IEEEbiographynophoto}
\vspace{-10 mm}
\begin{IEEEbiographynophoto}
{Jianwei Niu} (Senior Member, IEEE) received the Ph.D. degree in computer science and engineering from Beihang University, Beijing, China.
\par
\hspace{-1em} He is currently a Professor of Computer Science and Technology with Beihang University, Beijing, China. His research interests include robot operating systems, mobile computing, image processing, and natural language processing.
\end{IEEEbiographynophoto}
\vspace{-10 mm}
\begin{IEEEbiographynophoto}
{Mingliang Xu} received the Ph.D. degree in computer science and technology from the State Key Laboratory of CAD\&CG, Zhejiang University, Hangzhou, China.
\par
\hspace{-1em} He is currently a Full Professor with the School of Computer and Artificial Intelligence, Zhengzhou University, Zhengzhou, China, and also the Director of Engineering Research Center of Intelligent Swarm Systems, Ministry of Education, and the Vice General Secretary of ACM SIGAI China. He has authored more than 60 journal and conference articles. His research interests include computer graphics, multimedia, and artificial intelligence. 
\end{IEEEbiographynophoto}

\end{document}